\documentclass[lettersize,journal]{IEEEtran}

\usepackage{textcomp}
\usepackage{stfloats}
\usepackage{url}
\usepackage{verbatim}
\usepackage{graphicx}
\usepackage{cite}

\usepackage[utf8]{inputenc} 
\usepackage[T1]{fontenc}    
\usepackage{hyperref}       
\usepackage{url}            
\usepackage{booktabs}       
\usepackage{amsfonts}       
\usepackage{nicefrac}       
\usepackage{microtype}      
\usepackage{xcolor}         
\usepackage{amsmath}

\usepackage{colortbl} 
\usepackage{arydshln}
\usepackage[dvipsnames]{xcolor}
\usepackage{pifont} 
\usepackage{tikz} 
\definecolor{tabfirst}{rgb}{1, 0.75, 0.7}
\definecolor{tabsecond}{rgb}{1, 0.83, 0.7}
\definecolor{tabthird}{rgb}{1, 0.96, 0.7}

\usepackage{multicol}
\usepackage{multirow}
\usepackage{tabularx}
\usepackage{adjustbox}

\usepackage{graphicx}
\usepackage{subcaption}

\title{RoGS: Adaptive Meshgrid Gaussian for Large-Scale Road Surface Mapping}

\author{%
  Tianchen Deng, Zhiheng Feng, Wenhua Wu, Ziming Li, Chang Nie, Siting Zhu, Hesheng Wang,~\IEEEmembership{Senior Member,~IEEE}
\thanks{Tianchen Deng, Zhiheng Feng, Wenhua Wu, Ziming Li, Chang Nie, Siting Zhu, Hesheng are with School of Automation and Intelligent Sensing, Shanghai Jiao Tong university and State Key Laboratory of Avionics Integration and Aviation System-of-Systems Synthesis, Shanghai Key Laboratory of Navigation and Location Based Services, Shanghai 200240, China. This work was supported by National Key R\&D Program of China (Grant No.2024YFB4708900). It was also supported in part by the Natural Science Foundation of
China under Grant 62225309,U24A20278, 62361166632. The first two authors contribute equal to this paper.  (*corresponding author: wanghesheng@sjtu.edu.cn)
	}
}
\begin{document}

\maketitle

\begin{abstract}
  Road surface mapping plays a crucial role in autonomous driving, supporting high-definition map generation, lane-level perception, and automatic road annotation. Recent mesh-based road surface reconstruction methods have shown promising results, but they still suffer from limited reconstruction quality and high optimization cost, especially in large-scale driving scenarios. To address these limitations, we propose ROADGS-T, a robust and efficient large-scale road surface mapping framework based on adaptive meshgrid Gaussian representation. Specifically, we model the road surface by placing 2D Gaussian surfels on a meshgrid, where each surfel explicitly stores color, semantic, and geometric information. Compared with conventional mesh-based representations and 3D Gaussian primitives, the proposed meshgrid Gaussian representation better matches the thin-surface property of roads while significantly reducing redundant primitives and overlap during optimization. To further improve representation efficiency and structural fidelity, we introduce a road-structure-aware adaptive meshgrid strategy, which allocates denser Gaussian surfels to geometrically or semantically complex regions, such as lane markings, road boundaries, and height discontinuities, while maintaining a compact representation in flat road areas. Moreover, instead of relying on a single nearest vehicle pose, we design a trajectory-consistency-guided pose-robust refinement strategy, which estimates local surface priors from multiple neighboring poses and adaptively weights pose-guided height regularization according to their geometric consistency. This improves robustness to practical pose noise and local trajectory perturbations. Extensive experiments on KITTI and nuScenes demonstrate that RoGS achieves high-quality RGB, semantic, and elevation map reconstruction in various challenging real-world scenes, while substantially accelerating large-scale road surface mapping compared with existing mesh-based methods.
  The code will be open-sourced on \href{https://github.com/fzhiheng/RoGS}{https://github.com/fzhiheng/RoGS}.

\end{abstract}


\section{Introduction}
\label{sec:intro}

\begin{figure*}
\centering
\includegraphics[width=0.95\linewidth]{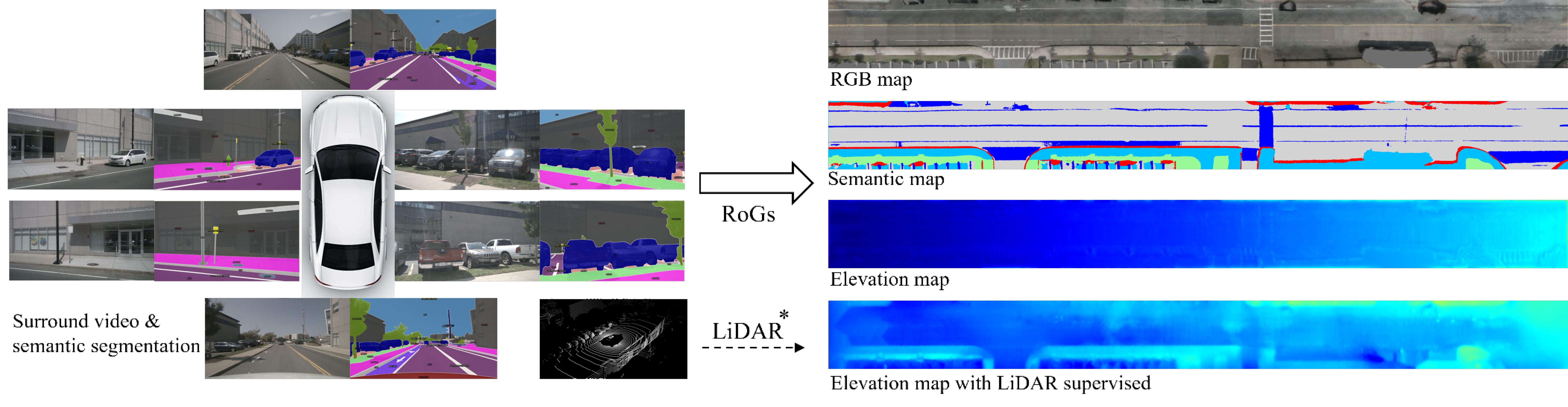}
\caption{We propose RoGS, a robust and efficient large-scale road surface mapping framework based on adaptive meshgrid Gaussian representation. Given surround-view videos, semantic segmentation results, and vehicle poses, RoGS reconstructs RGB maps, semantic maps, and elevation maps of road surfaces.}
\label{fig:pipeline}
\end{figure*}

The rapid development of autonomous driving technology has created an increasing demand for accurate, efficient, and scalable road surface mapping. Road surface reconstruction aims to recover the geometric, visual, and semantic information of drivable areas from onboard sensor data, including road regions, lane lines, road markings, curbs, and other traffic-related surface structures. Such reconstructed road surface maps can be directly used for high-definition map generation and automatic data annotation~\cite{wu2024emie, mei2024rome}. They can also provide dense geometric and semantic priors for downstream autonomous driving tasks, such as BEV perception, road topology understanding, and safe motion planning~\cite{zhao2024roadbev, zhao2023rsrd}. Therefore, building high-quality road surface maps from vehicle-collected data is an important problem for intelligent transportation systems.

Traditional 3D reconstruction methods mainly focus on recovering the overall geometric structure of scenes. Structure-from-Motion (SfM)~\cite{ozyecsil2017survey} and Multi-View Stereo (MVS) methods~\cite{zhou2021dp} can reconstruct sparse or semi-dense point clouds from multi-view images. However, road surfaces usually contain large textureless regions, repetitive lane markings, illumination changes, and occlusions caused by dynamic objects, which often lead to holes, noise, and incomplete reconstruction. Neural Radiance Field (NeRF)-based methods~\cite{tancik2022block, turki2022mega, barron2022mip, song2023sc, zhu2023sni} have achieved impressive novel-view synthesis quality, but their implicit scene representations make it difficult to explicitly recover accurate road geometry, especially in large-scale outdoor driving scenarios. Recently, 3D Gaussian Splatting~\cite{kerbl20233d, chen2024survey} has significantly improved rendering efficiency through explicit Gaussian primitives. Nevertheless, most Gaussian-based methods are designed for photorealistic rendering rather than road surface geometry recovery, and they do not fully exploit the thin-surface property and regular spatial structure of road surfaces.

To address road surface reconstruction more directly, recent methods have introduced explicit meshes or hybrid explicit-implicit representations for large-scale road modeling~\cite{wu2024emie, mei2024rome}. These methods have demonstrated promising results in reconstructing RGB and semantic road maps. However, their optimization is still time-consuming, and the reconstruction quality remains limited in complex driving scenes. More importantly, existing road surface reconstruction methods usually rely on fixed mesh structures or dense surface elements, which may introduce redundant computation in flat road regions while being insufficiently detailed around lane markings, curbs, road boundaries, and height discontinuities. In addition, vehicle poses are commonly used as important geometric priors for road surface initialization. However, practical vehicle poses inevitably contain local noise and trajectory perturbations, which may degrade the height initialization and subsequent surface optimization. These issues limit the practical deployment of existing methods in large-scale autonomous driving applications.

In this paper, we propose RoGS, a robust and efficient large-scale road surface mapping framework with adaptive meshgrid Gaussian representation. The core idea is to model the road surface using mesh-distributed 2D Gaussian surfels. Specifically, we place Gaussian surfels on the vertices of a meshgrid, where each surfel explicitly stores color, semantic, and geometric information. Since road surfaces are thin structures without physical thickness, 2D Gaussian surfels are more consistent with the actual geometry of road surfaces than conventional 3D Gaussian spheres. Meanwhile, the meshgrid layout provides a compact and structured road representation, covering large-scale driving areas with fewer Gaussian surfels and reducing unnecessary overlap during optimization.

Building upon this representation, we further introduce a road-structure-aware adaptive meshgrid strategy. Instead of using a uniformly dense grid for the entire road surface, RoGS adaptively allocates Gaussian surfels according to local road complexity. For flat and textureless road regions, a compact grid is sufficient to represent the surface. For geometrically or semantically complex regions, such as lane markings, road boundaries, curbs, crosswalks, and height discontinuities, denser Gaussian surfels are introduced to preserve fine-grained structures. This adaptive design improves representation efficiency while maintaining high reconstruction fidelity for traffic-critical road elements.

Moreover, we propose a trajectory-consistency-guided pose-robust refinement strategy to reduce the influence of unreliable local pose priors. In the original pose-based initialization, the height and rotation of each Gaussian surfel are initialized using the nearest vehicle pose under the assumption that the vehicle pose is locally parallel to the road surface. Although effective, this strategy may be sensitive to pose noise or local trajectory perturbations. To improve robustness, RoGS estimates local surface priors from multiple neighboring vehicle poses and measures their geometric consistency. When neighboring poses provide consistent height predictions, the pose-guided prior is assigned a higher confidence. Otherwise, the influence of the pose prior is weakened, allowing the rendering, semantic, smoothness, and optional LiDAR elevation supervision to refine the road surface more flexibly. This design improves the stability of road surface optimization under practical pose uncertainty without requiring explicit pose covariance.

After initialization and adaptive refinement, RoGS optimizes the road surface through differentiable Gaussian rendering. The rendered RGB images and semantic maps are supervised by camera images and semantic segmentation results. In addition, LiDAR point clouds can be optionally used to supervise the elevation of Gaussian surfels when available. Finally, RoGS reconstructs large-scale road surface maps with RGB, semantic, and elevation information. Extensive experiments on the KITTI and nuScenes datasets demonstrate that the proposed method achieves efficient and high-quality road surface reconstruction in various challenging real-world driving scenes. In particular, compared with existing mesh-based road reconstruction methods, RoGS significantly improves optimization efficiency while preserving fine road structures and producing accurate semantic and elevation maps.

The main contributions of this paper are summarized as follows:

\begin{itemize}
\item We propose RoGS, a robust and efficient large-scale road surface mapping framework based on meshgrid Gaussian representation. The proposed 2D Gaussian surfel representation explicitly models color, semantics, and geometry while conforming to the thin-surface property of road surfaces.

\item We introduce a road-structure-aware adaptive meshgrid strategy, which allocates Gaussian surfels according to local geometric and semantic complexity. This design improves representation efficiency and preserves fine-grained traffic-critical structures, such as lane markings, road boundaries, curbs, and height discontinuities.

\item We propose a trajectory-consistency-guided pose-robust refinement strategy. By aggregating local surface priors from multiple neighboring vehicle poses and adaptively weighting the pose-guided height regularization according to their geometric consistency, RoGS reduces the influence of local pose noise without requiring explicit pose covariance.

\item Extensive experiments on the KITTI and nuScenes datasets demonstrate that RoGS achieves efficient and high-quality road surface reconstruction. Compared with existing mesh-based methods, the proposed framework substantially improves optimization efficiency, achieving speedups of \textbf{53$\times$} and \textbf{27$\times$} when iterating one epoch and two epochs, respectively, while producing accurate RGB, semantic, and elevation maps.
\end{itemize}

\begin{figure*}[htp]
  \centering
  \includegraphics[width=0.9\linewidth]{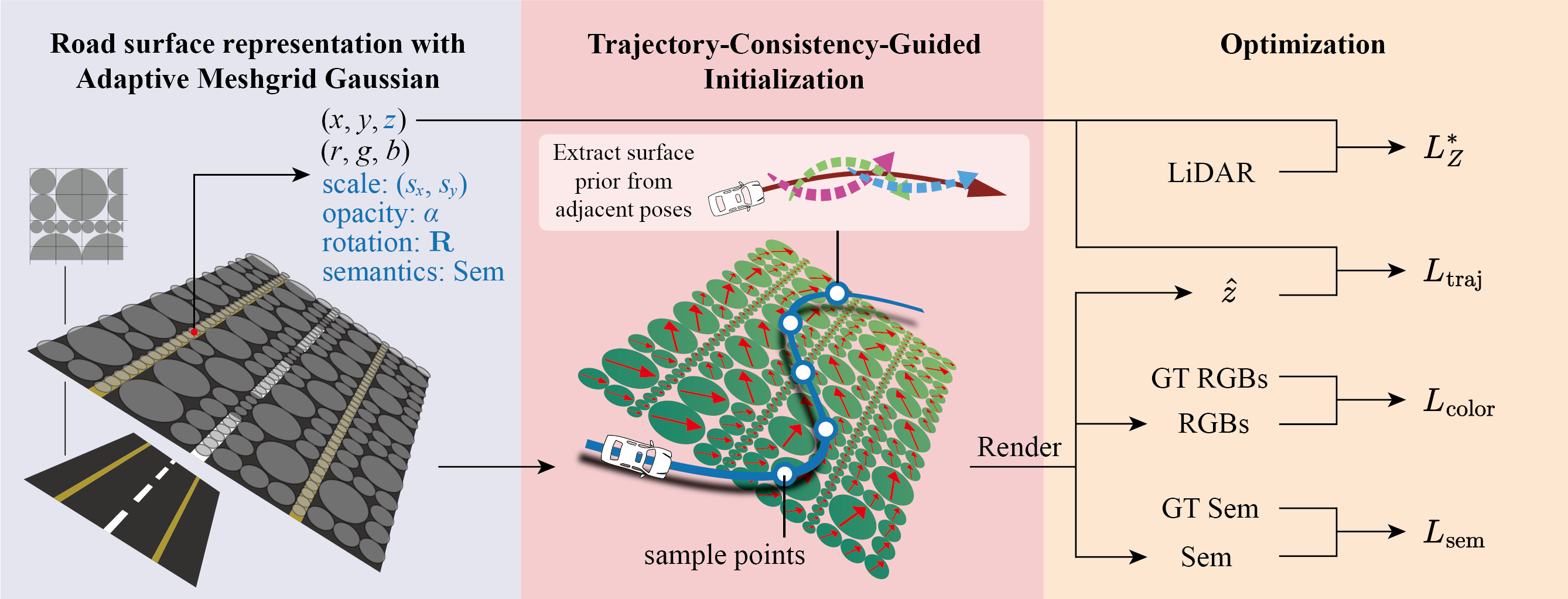}
  \caption{Overview of RoGS. The left side shows the road representation with meshgrid Gaussian. The blue curve indicates the vehicle trajectory. Mesh-distributed Gaussian surfels are used to represent the road surface. Each surfel stores position, scale, rotation, color, opacity, and semantic information. The learnable parameters are indicated in \textcolor[HTML]{0070C0}{blue}. The middle demonstrates the pose-based initialization. For each Gaussian surfel, its elevation (z) and rotation are initialized using the nearest vehicle pose. Finally, the rendering is supervised by camera images and semantic segmentation results. Additionally, to improve the reconstruction quality, LiDAR point clouds can be introduced to supervise the elevation. $L^{*}$ indicates an optional loss.}
  \label{fig:overview}
\end{figure*}

\section{Related Works}
\label{sec:relate}

\subsection{Classical 3D Reconstruction}

Reconstructing 3D scenes from multi-view images is a fundamental task in computer vision, robotics, and autonomous driving. Traditional 3D reconstruction methods usually represent scenes as point clouds or meshes, and have achieved remarkable progress over the past decades~\cite{ozyecsil2017survey}. Structure-from-Motion (SfM) methods recover camera poses and sparse 3D structures by extracting image features, matching correspondences, triangulating 3D points, and performing bundle adjustment. COLMAP~\cite{Schonberger2016structure} is a representative incremental SfM system and has been widely used for large-scale scene reconstruction~\cite{snavely2006photo, snavely2011scene, wu2013towards, moulon2013adaptive}. However, SfM-based methods mainly rely on sparse feature correspondences, and the reconstructed point clouds are often incomplete in textureless or repetitive regions~\cite{liu2023regformer}.

Multi-View Stereo (MVS) methods further densify the reconstructed geometry by estimating dense correspondences across multiple views~\cite{schonberger2016pixelwise, zhou2021dp, agarwal2011building}. Although MVS methods can generate denser scene structures than SfM, they still heavily depend on reliable image matching. In road surface reconstruction, large areas of asphalt are weakly textured, lane markings are often repetitive, and illumination changes or dynamic objects may introduce severe matching ambiguity. As a result, classical reconstruction methods tend to produce holes, noisy points, and incomplete geometry on road surfaces. These limitations make it difficult to directly apply general-purpose 3D reconstruction pipelines to large-scale road surface mapping in autonomous driving scenarios.

\subsection{Neural Implicit and Gaussian Scene Representations}

Neural scene representations have recently attracted increasing attention for 3D reconstruction and novel-view synthesis. Neural Radiance Fields (NeRF)~\cite{mildenhall2021nerf} represent a scene as a continuous radiance field using Multi-Layer Perceptrons (MLPs), achieving high-quality view synthesis through differentiable volume rendering. Mip-NeRF~\cite{barron2021mip} introduces integrated positional encoding based on conical frustums to improve anti-aliasing. Some methods combine neural networks with grid-based representations to improve efficiency and scalability~\cite{wang2023co, wu2025dvn}. For large-scale scenes, Mip-NeRF 360~\cite{barron2022mip}, Mega-NeRF~\cite{turki2022mega}, Block-NeRF~\cite{tancik2022block}, BungeeNeRF~\cite{xiangli2022bungeenerf}, and ProSGNeRF~\cite{deng2023prosgnerf} extend neural rendering to unbounded or city-scale scenarios. Several methods further explore neural representations for urban scenes and autonomous driving~\cite{turki2023suds, xie2023s, song2023sc, zhu2023sni}. However, NeRF-based methods usually require expensive training and rendering, and their implicit representations make it difficult to explicitly recover accurate road surface geometry and semantic maps.

The emergence of 3D Gaussian Splatting (3DGS)~\cite{kerbl20233d, chen2024survey} significantly improves rendering efficiency by representing scenes with explicit Gaussian primitives. Compared with implicit neural fields, 3DGS provides a more explicit and editable representation, making it attractive for robotic perception, mapping, and autonomous driving applications. Recent surveys have discussed the evolution of 3D scene representations from traditional geometry to neural fields, Gaussian representations, and foundation models, highlighting the importance of choosing task-oriented representations for robotics and embodied systems~\cite{deng2025best3dscenerepresentation}. Although standard 3DGS is highly effective for photorealistic rendering, it is mainly designed for novel-view synthesis rather than structured road surface reconstruction. Directly using 3D Gaussian spheres to model road surfaces may introduce redundant primitives and unnecessary thickness, while failing to fully exploit the planar and thin-surface characteristics of roads. In contrast, our method uses mesh-distributed 2D Gaussian surfels to explicitly model road surfaces, which better matches the physical structure of roads and improves optimization efficiency.

\subsection{Gaussian-based SLAM and Reconstruction}

Beyond offline novel-view synthesis, Gaussian representations have also been widely explored for SLAM, dense mapping, and robotic reconstruction. Gaussian-based SLAM methods usually integrate differentiable Gaussian rendering into online tracking and mapping systems, enabling dense reconstruction, real-time rendering, and explicit map optimization. SemGauss-SLAM~\cite{zhu2024semgauss} incorporates semantic information into Gaussian maps for dense semantic SLAM. Compact Gaussian SLAM methods reduce redundant Gaussian primitives to improve memory efficiency and runtime performance~\cite{compactslam}. Voxel-based progressive Gaussian SLAM further organizes Gaussian primitives with voxel structures and progressive optimization, improving scalability in large-scale scenes~\cite{deng2025vpgs}. Other recent works also explore Gaussian representations for dynamic scenes, point cloud interpolation, and neural mapping~\cite{jiang2024neurogauss4dpci}. These works demonstrate that Gaussian representations are promising for robotic mapping because of their explicit geometry, differentiable rendering, and efficient optimization.

\subsection{Road Surface Reconstruction}

Road surface reconstruction directly focuses on recovering dense road geometry and appearance from vehicle-mounted sensors. Early methods reconstruct road surfaces using stereo vision, mobile LiDAR, or multi-sensor fusion~\cite{fan2018road, yu20073d, brunken2020road, guo2015automatic, fan2021rethinking}. These methods can generate useful local road profiles or surface models, but they are usually limited to small-scale scenes and often require dense depth measurements or carefully designed geometric assumptions. Recently, road surface reconstruction has become increasingly important for autonomous driving, since reconstructed road maps can support HD map generation, road annotation, lane-level perception, and safety-oriented driving analysis. RoadBEV~\cite{zhao2024roadbev} and RSRD~\cite{zhao2023rsrd} further emphasize the importance of road surface reconstruction for BEV perception and safe autonomous driving.

For large-scale road surface reconstruction, RoMe~\cite{mei2024rome} represents the road surface using an explicit mesh and optimizes vertex colors and semantics through differentiable rendering. EMIE-MAP~\cite{wu2024emie} further introduces implicit color encoding to address the inconsistent exposure and illumination problem in surround-view cameras. These methods achieve promising results in reconstructing road RGB and semantic maps. However, mesh-based rendering optimization remains time-consuming, and fixed mesh structures may either introduce redundant computation in simple road regions or lose details in complex regions such as lane markings, curbs, crosswalks, and road boundaries. Moreover, existing methods usually rely on vehicle poses or point clouds for initialization, but the influence of local pose noise and trajectory inconsistency is not sufficiently considered.

Different from these methods, we propose a robust and efficient road surface mapping framework based on adaptive meshgrid Gaussian representation. By placing 2D Gaussian surfels on a structured meshgrid, our method explicitly models color, semantics, and elevation while maintaining a compact road representation. Furthermore, the proposed road-structure-aware adaptive meshgrid strategy allocates more Gaussian surfels to traffic-critical and geometrically complex areas, improving local reconstruction fidelity without introducing unnecessary redundancy. In addition, the trajectory-consistency-guided pose-robust refinement strategy reduces the influence of unreliable local pose priors by aggregating surface cues from multiple neighboring vehicle poses. As a result, our method achieves efficient, accurate, and robust large-scale road surface mapping for autonomous driving.

\section{Method}

The overview of RoGs is shown \ref{fig:overview}, which consists of three main components. First, we use meshgrid Gaussian to represent the real road surface . The Gaussian surfels store position, scale, rotation, color, opacity, and semantic information. Then, for each Gaussian surfel, its elevation (z) and rotation are initialized using the nearest vehicle pose. Finally, the rendered RGB and semantics are supervised by the camera images and semantic segmentation results. Additionally, LiDAR point clouds can be introduced to supervise the elevation of surfels to improve the reconstruction quality.

\begin{figure}[htbp]
  \centering
    \begin{subfigure}{0.25\linewidth} 
      \includegraphics[width=\linewidth]{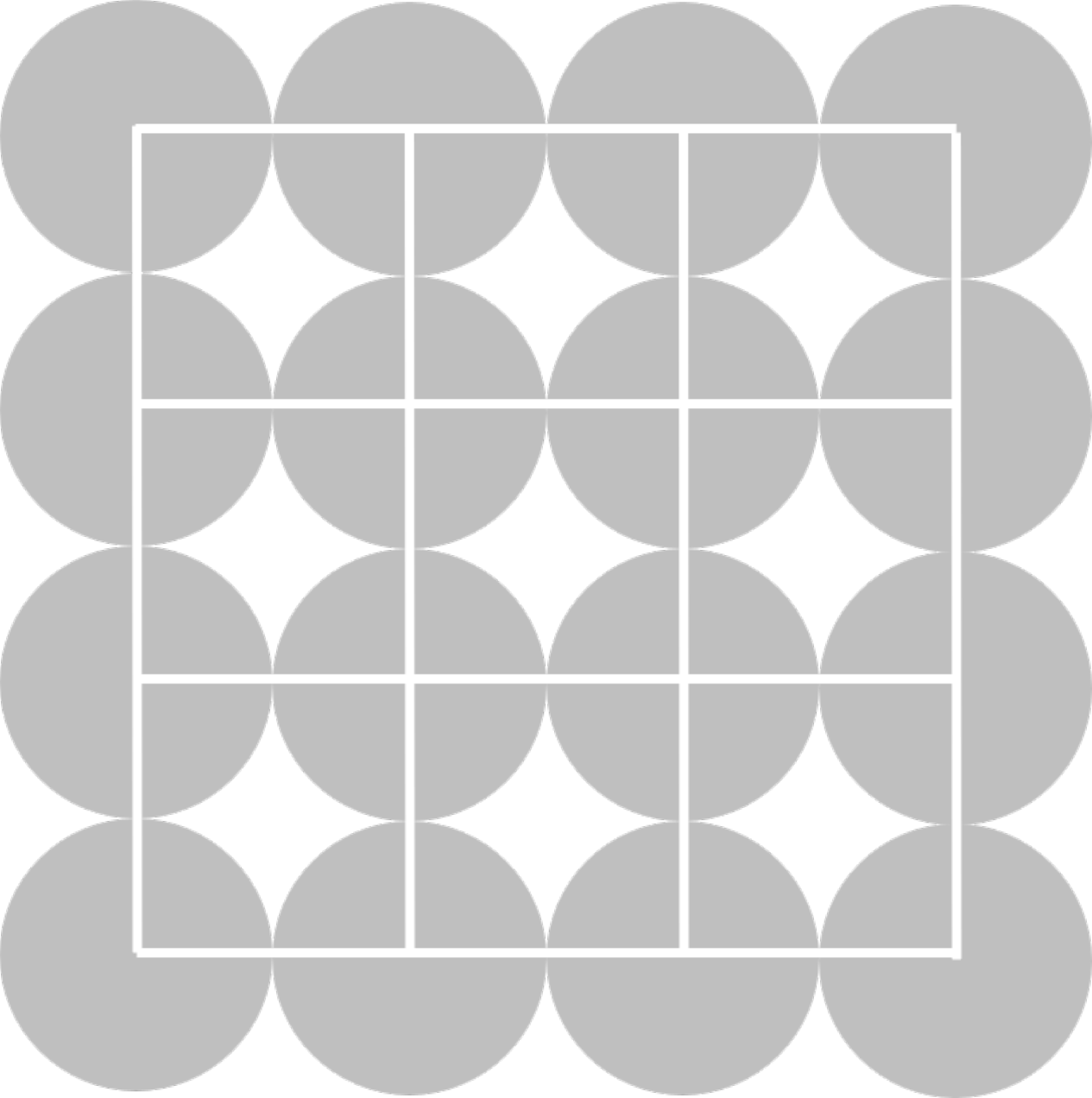}
        \caption{Layout-1}
        \label{fig:layout1}
    \end{subfigure}  \qquad \begin{subfigure}{0.27\linewidth}
      \includegraphics[width=\linewidth]{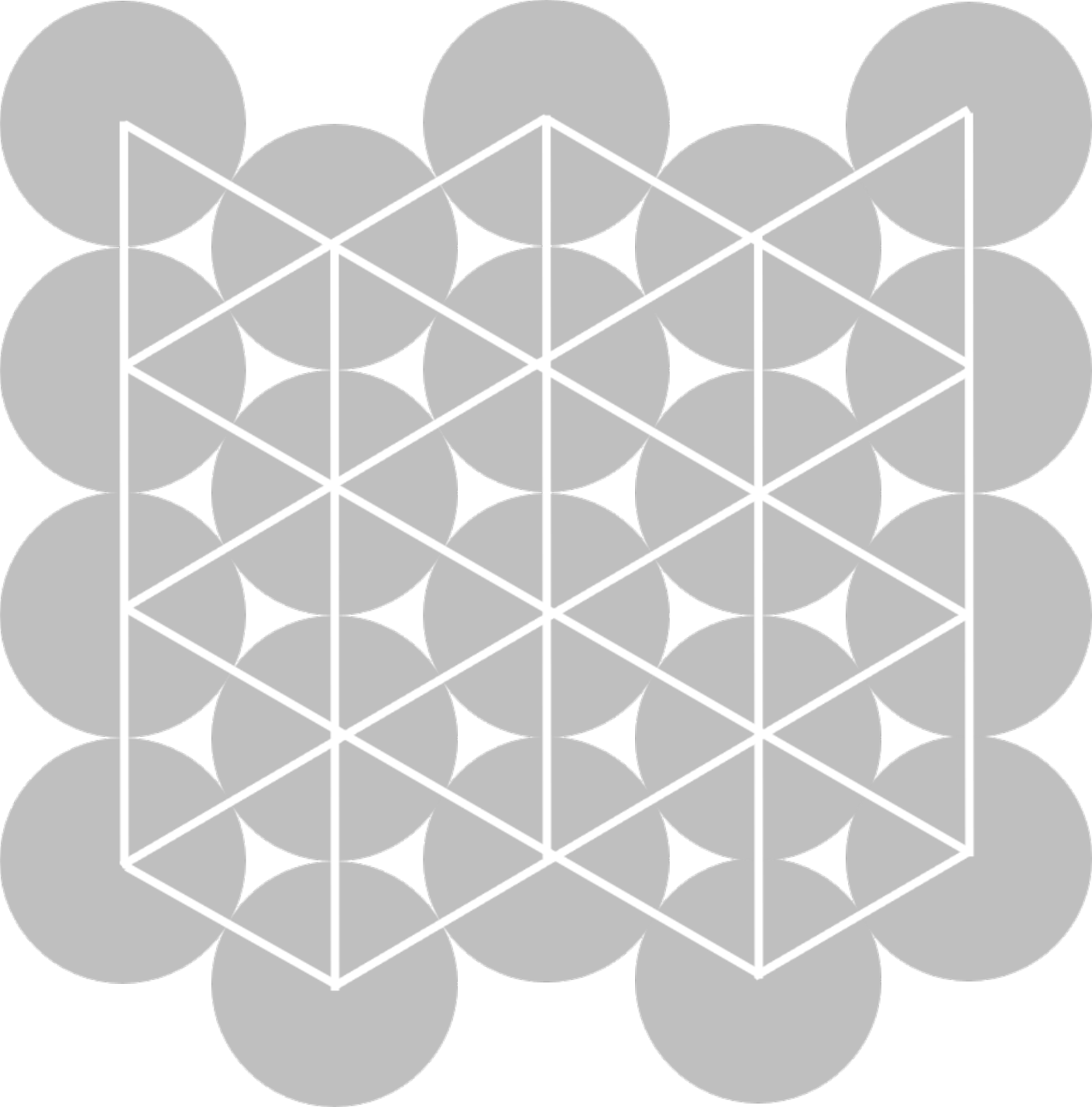}
        \caption{Layout-2}
        \label{fig:layout2}
    \end{subfigure}
\caption{Comparison of the two layouts. Layout-1 covers the entire road with fewer Gaussian surfels and reduces the overlap between Gaussian surfels during training. Therefore, Layout-1 is used to place Gaussian surfels.}
\label{fig:meshgrid}
\end{figure}

\begin{figure}[tbp]
  \centering
  \includegraphics[width=\linewidth]{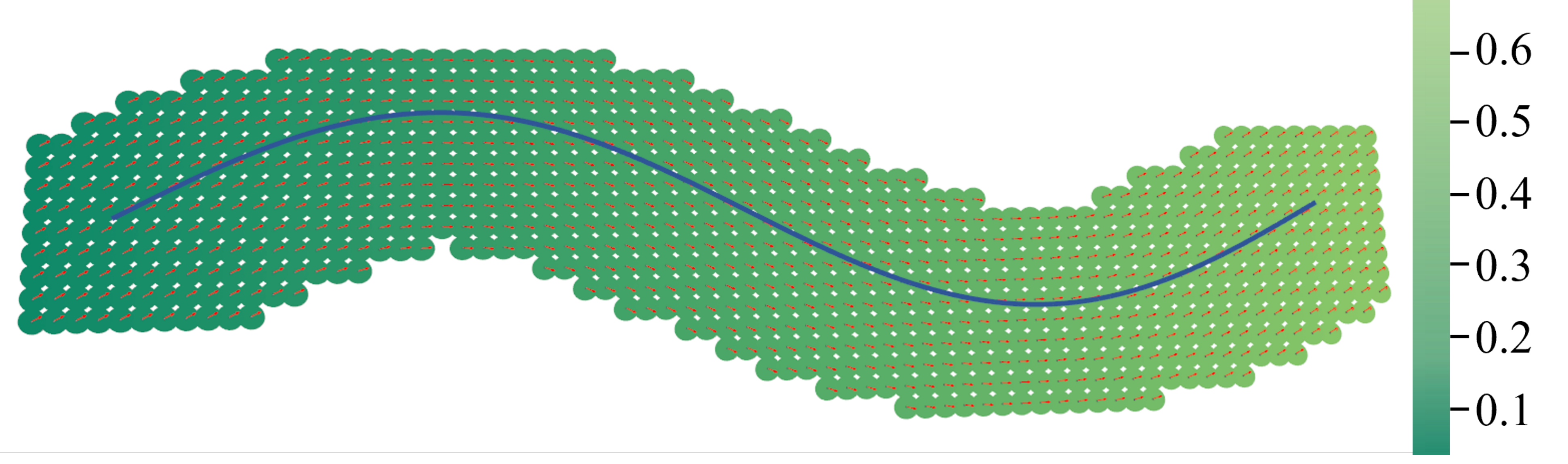}
  \caption{Pose-based initialization. For each Gaussian surfel, its elevation and rotation are initialized using the nearest vehicle pose. The blue curve represents the vehicle trajectory, and the red arrows represent the x-axis.}
  \label{fig:road}
\end{figure}

\begin{figure}[ht]
  \centering
  \includegraphics[width=\linewidth]{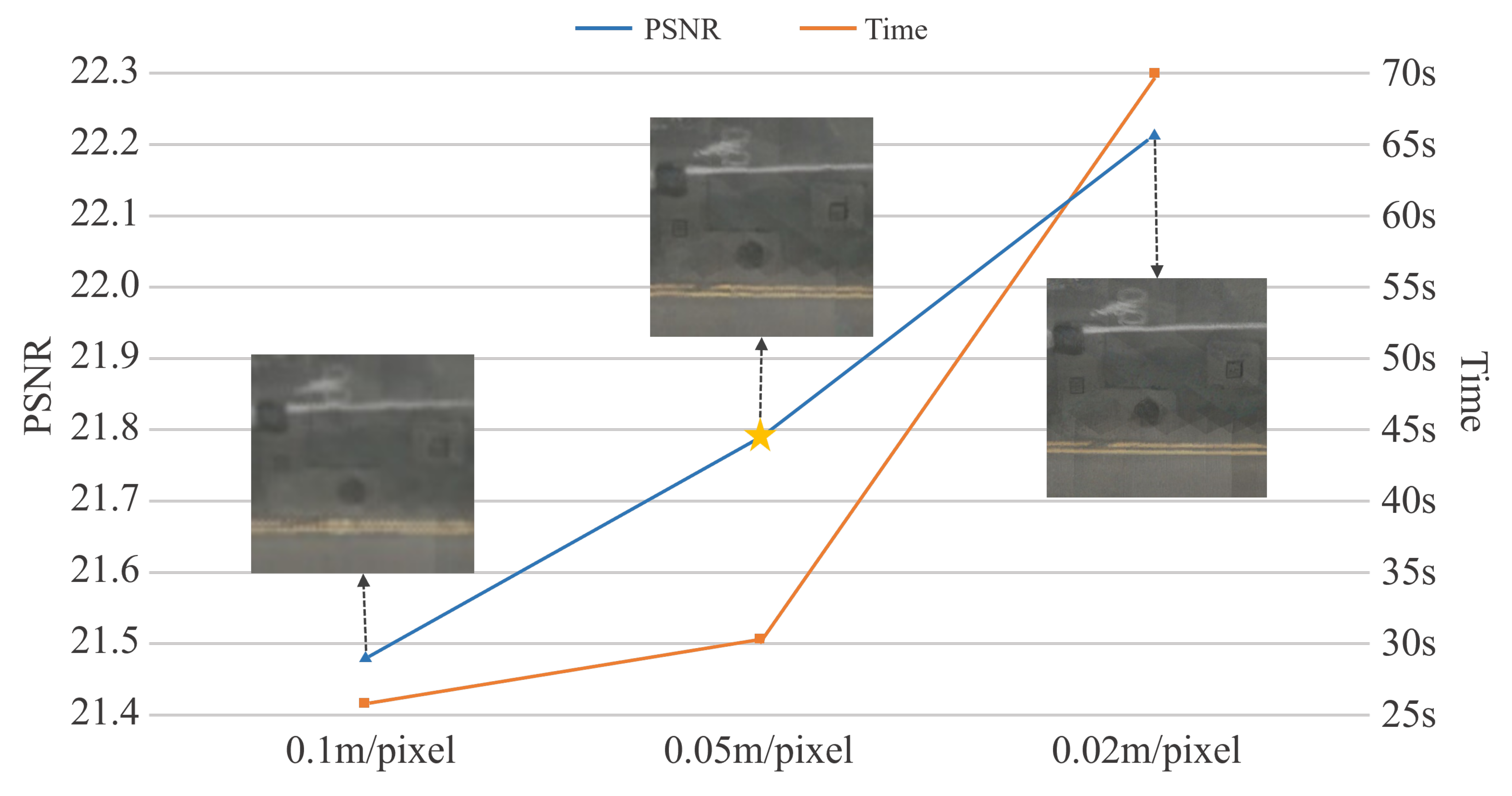}
  \caption{Results at different resolutions. 0.05m/pixel resolution achieves realistic reconstruction with improved training speed.}
  \label{fig:resolution}
\end{figure}

\begin{figure*}
     \centering
  \includegraphics[width=\linewidth]{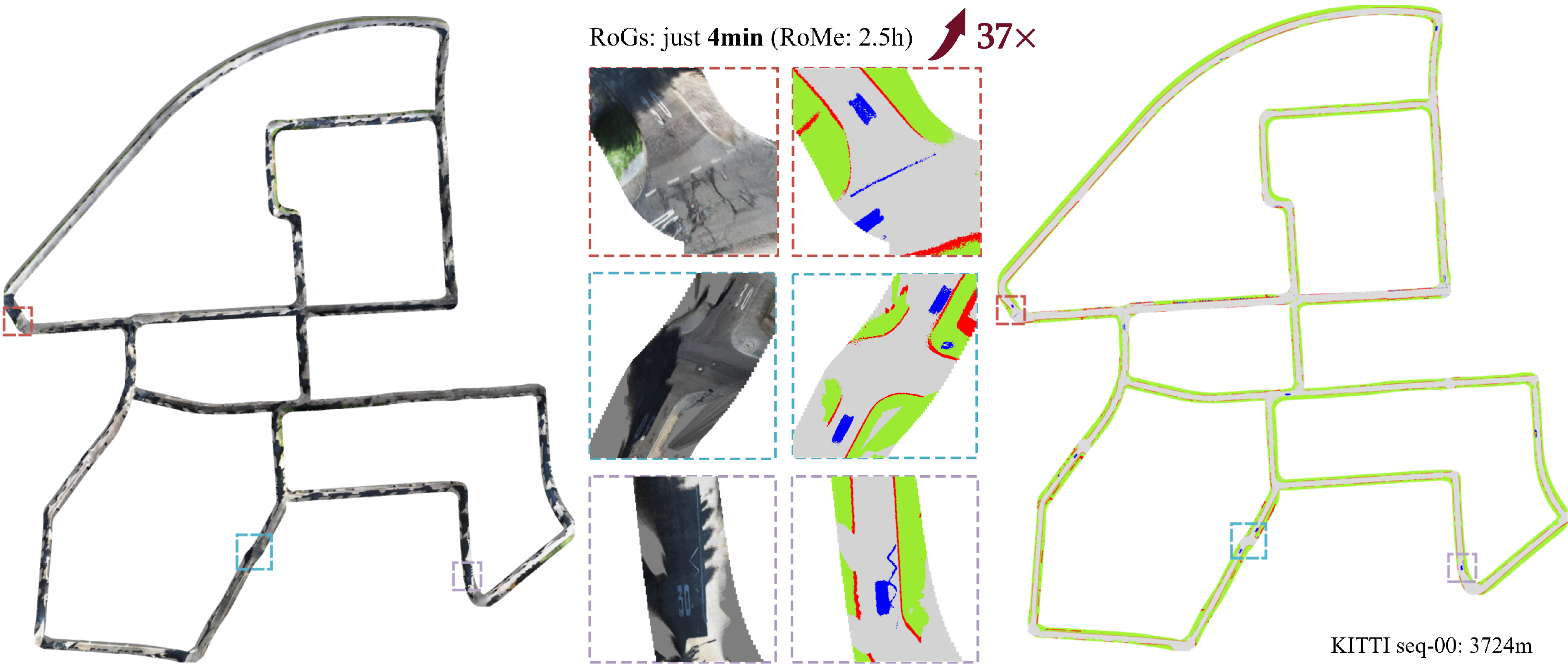}
  \captionof{figure}{Road surface reconstruction results (KITTI odometry sequence-00) using our proposed RoGs, covering a trajectory length of roughly 3724m. We reconstruct the complete scene in just \textbf{4 minutes}, while RoMe~\cite{mei2024rome} takes 2.5 hours.}
  \label{fig:kitti}
\end{figure*}

\begin{figure*}[t]
  \centering
  \includegraphics[width=\linewidth]{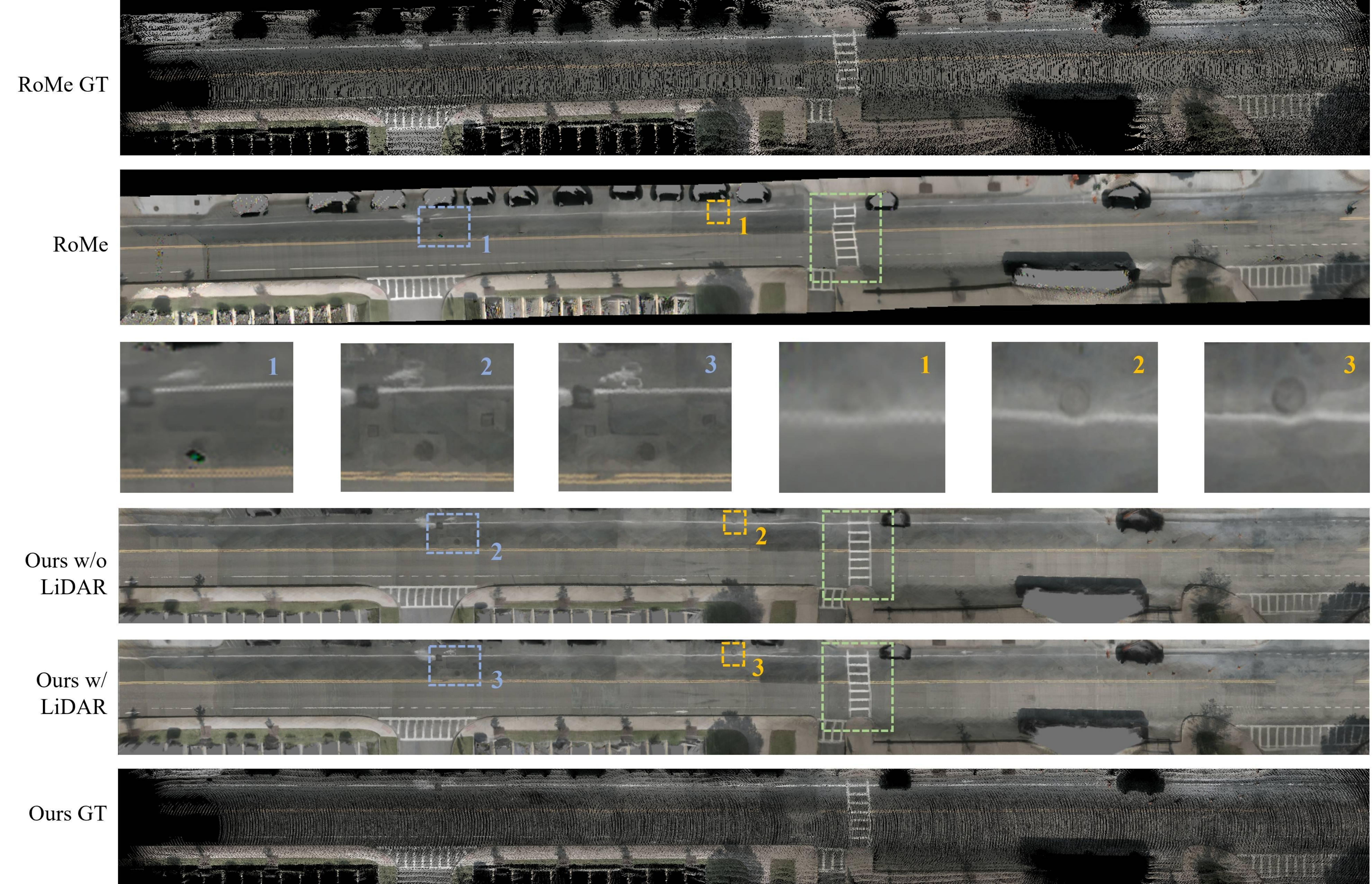}
  \caption{RGB results of road surface reconstruction for Scene-0655. Because RoMe is in a different reference to ours, the ground truth needs to be transformed into the corresponding reference for rendering. In the local detail images, our method reconstructs both the ground bike markings and lane lines more clearly, which shows that our method is able to reconstruct finer details. In addition, when height is supervised using LiDAR, the quality and detail of the rendering are further improved.}
  \label{fig:0655_rgb}
\end{figure*}

\begin{figure*}[tb]
  \centering
  \includegraphics[width=\linewidth]{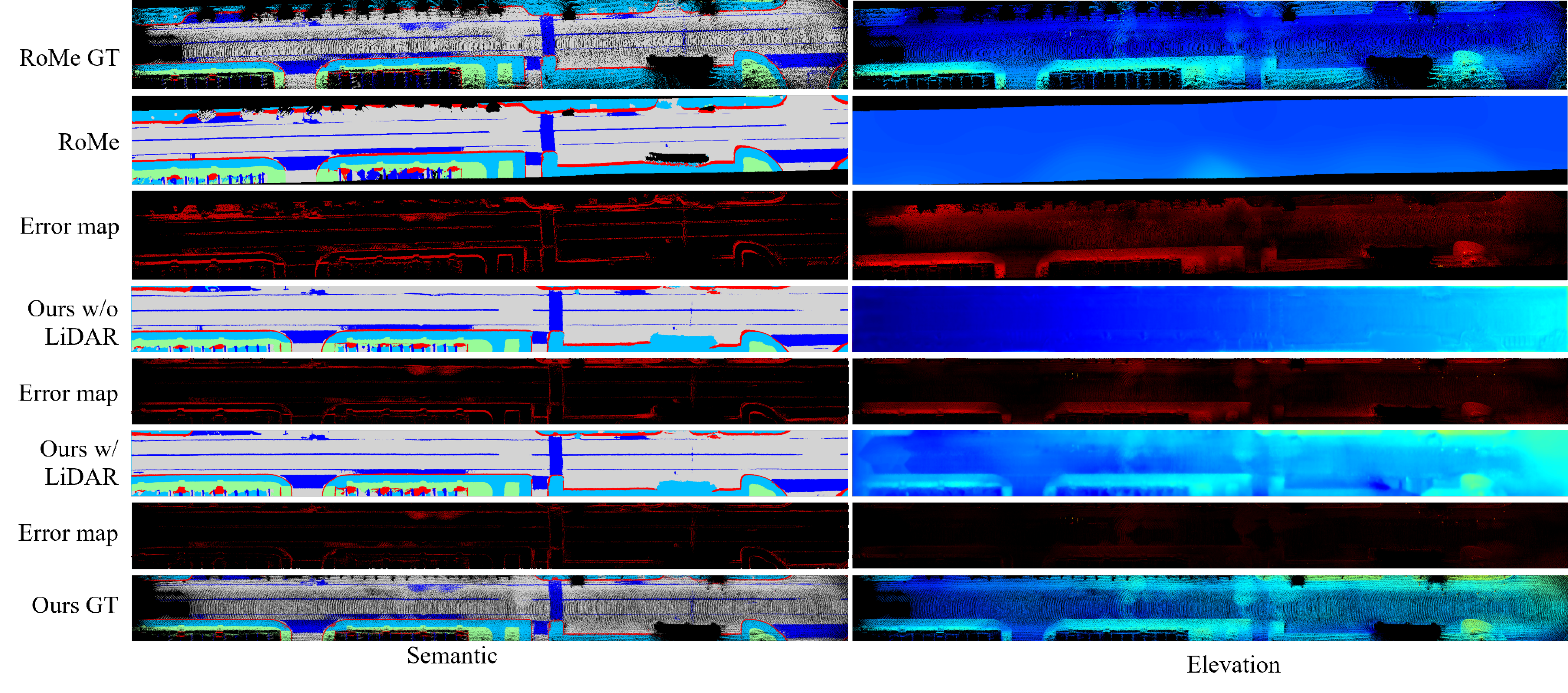}
  \caption{Semantic and elevation results of road surface reconstruction for Scene-0655. For the elevation error map, brighter colors indicate larger errors. For the semantic error map, incorrect pixels are shown in red. The error maps show that our method has fewer errors. In addition, when neither LiDAR point cloud is used, the reconstructed elevation of our method restores the height of road edges and zebra crossings. In contrast, the elevation reconstructed by RoMe is too smooth and does not correspond to real-world physical road surfaces.}
  \label{fig:0655_sem}
\end{figure*}

\subsection{Representation with Adaptive Meshgrid Gaussian}
\label{sec:meshgrid}

In this section, we introduce the proposed adaptive meshgrid Gaussian representation for road surface mapping. Different from general 3D scene reconstruction methods that model the entire scene with volumetric primitives, RoGS focuses on the thin road surface and represents it using mesh-distributed 2D Gaussian surfels. Each surfel explicitly stores geometric, appearance, opacity, and semantic information, enabling the reconstruction of RGB maps, semantic maps, and elevation maps in a unified representation.

\textbf{2D Gaussian Surfel:} In 3DGS~\cite{kerbl20233d}, a Gaussian primitive in the world space is modeled by a 3D covariance matrix $\mathbf{\Sigma}$ and its center coordinate $\mathbf{p}$:
\begin{equation}
G(\mathbf{p})=e^{-\frac{1}{2}\mathbf{p}^{T}\mathbf{\Sigma}^{-1}\mathbf{p}}.
\end{equation}
The 3D covariance matrix $\mathbf{\Sigma}$ is determined by the rotation matrix $\mathbf{R}$ and the scale vector $\mathbf{s}=(s_x,s_y,s_z)^T$. For convenience, the scale is represented as a diagonal matrix $\mathbf{S}=\operatorname{Diag}(s_x,s_y,s_z)$. The covariance matrix can be obtained as:
\begin{equation}
\mathbf{\Sigma}=\mathbf{R}\mathbf{S}\mathbf{S}^{T}\mathbf{R}^{T}.
\end{equation}

Similar to previous surface-oriented Gaussian representations~\cite{dai2024high, huang20242d}, we set the scale along the normal direction $s_z$ to 0 to obtain a 2D Gaussian surfel. Compared with 3D Gaussian spheres, 2D Gaussian surfels are more consistent with the physical property of road surfaces, since roads are thin surfaces without explicit thickness. To represent road appearance and semantics, we further parameterize the color, opacity, and semantic attributes of each Gaussian surfel. Finally, a Gaussian surfel is explicitly parameterized as:
\begin{equation}
\Theta_i={(x_i,y_i,z_i),(r_i,g_i,b_i),(s_{x,i},s_{y,i}),\alpha_i,\mathbf{R}_i,\mathbf{Sem}_i},
\end{equation}
where $(x_i,y_i,z_i)$ denotes the surfel center, $(r_i,g_i,b_i)$ denotes the RGB color, $(s_{x,i},s_{y,i})$ denotes the surfel scale on the tangent plane, $\alpha_i$ denotes opacity, $\mathbf{R}_i$ denotes rotation, and $\mathbf{Sem}_i$ denotes the semantic vector. In practice, $\mathbf{Sem}_i$ is represented by a vector whose length is equal to the number of semantic categories.

\textbf{Base Meshgrid Layout:} To cover the road surface, we first construct a base meshgrid according to the vehicle trajectory. Specifically, the vehicle poses are projected onto the $xy$-plane, and the road range is expanded around the projected trajectory. The expanded road region is then divided into uniformly distributed square cells, where Gaussian surfels are placed on the vertices of the grid. As shown in Fig.~\ref{fig:layout1}, this layout covers the road surface with fewer Gaussian surfels than the alternative dense layout in Fig.~\ref{fig:layout2}. Since the scale of each Gaussian surfel is learnable, neighboring surfels may overlap during optimization. Excessive overlap can interfere with gradient-based optimization, while the adopted meshgrid layout reduces unnecessary neighboring relations and improves optimization efficiency.

\textbf{Road-Structure-Aware Adaptive Meshgrid:} Although a uniform meshgrid is efficient for representing large-scale road surfaces, different road regions have different structural complexity. Flat asphalt regions can be represented by relatively sparse surfels, while traffic-critical regions, such as lane markings, curbs, crosswalks, road boundaries, and height discontinuities, require denser surfels to preserve fine details. Therefore, we introduce a road-structure-aware adaptive meshgrid strategy to allocate Gaussian surfels according to local road complexity.

For each base grid cell $\mathcal{C}_m$, we compute a structure complexity score from visual, semantic, and geometric cues:
\begin{equation}
\kappa_m=\omega_g G_m+\omega_s S_m+\omega_z Z_m,
\end{equation}
where $G_m$ denotes the RGB gradient response, $S_m$ denotes the semantic boundary or semantic entropy response, $Z_m$ denotes the local height variation or height residual, and $\omega_g$, $\omega_s$, and $\omega_z$ are the corresponding weights. The RGB and semantic cues are obtained from the input images and semantic segmentation results after projection into the road coordinate system. When LiDAR is available, $Z_m$ is computed from the local elevation variation of the point cloud. Otherwise, it is computed from the height variation predicted by neighboring vehicle poses.

Based on the structure complexity score, the adaptive refinement indicator of each grid cell is defined as:
\begin{equation}
a_m=\mathbb{I}(\kappa_m>\tau_a),
\end{equation}
where $\mathbb{I}(\cdot)$ is the indicator function and $\tau_a$ is the adaptive refinement threshold. If $a_m=1$, the corresponding grid cell is subdivided into smaller cells, and additional Gaussian surfels are inserted at the refined vertices. Otherwise, the original coarse grid structure is retained. The final Gaussian surfel set is represented as:
\begin{equation}
\mathcal{V}=\mathcal{V}_{base}\cup\mathcal{V}_{adp},
\end{equation}
where $\mathcal{V}_{base}$ denotes the surfels from the base meshgrid and $\mathcal{V}_{adp}$ denotes the additional surfels generated by adaptive refinement. This adaptive design improves the representation capacity around complex road structures while avoiding redundant surfels in simple flat regions.

\subsection{Trajectory-Consistency-Guided Pose-Robust Initialization}
\label{sec:init}

After obtaining the initial Gaussian surfel coordinates on the $xy$-plane, their elevations are not aligned with the real road surface because the initial $z$ values are unknown. A natural prior in autonomous driving is that the vehicle pose is locally parallel to the road surface. Therefore, the road surface height and orientation can be initialized from vehicle poses. However, practical vehicle poses may contain local noise or trajectory perturbations. If each surfel only uses its nearest vehicle pose, the initialization may be sensitive to local pose errors. To improve robustness, we introduce a trajectory-consistency-guided pose-robust initialization strategy.

For a Gaussian surfel located at $(x_i,y_i)$, instead of only using its nearest vehicle pose, we search for its neighboring vehicle poses in the $xy$-plane:
\begin{equation}
\mathcal{N}_i={T_j}_{j=1}^{K},
\end{equation}
where $K$ is the number of neighboring poses. For each neighboring pose $T_j$, its position and rotation are denoted as $(x_j,y_j,z_j)$ and $\mathbf{R}_j=(\mathbf{n}_{j,1},\mathbf{n}_{j,2},\mathbf{n}_{j,3})$, respectively. The road normal estimated from this pose is $\mathbf{n}_{j,3}=(n_{j,x},n_{j,y},n_{j,z})^T$. Under the local road-plane assumption, this pose predicts the elevation of the surfel as:
\begin{equation}
z_i^{(j)}=z_j-\frac{1}{n_{j,z}}\left(n_{j,x}(x_i-x_j)+n_{j,y}(y_i-y_j)\right).
\end{equation}

The distance-based weight between the surfel and the neighboring pose is defined as:
\begin{equation}
w_{ij}=\exp\left(-\frac{|(x_i,y_i)-(x_j,y_j)|_2^2}{2\sigma_d^2}\right),
\end{equation}
where $\sigma_d$ controls the spatial influence range of neighboring poses. The pose-consensus elevation is then computed by weighted averaging:
\begin{equation}
\hat{z}_i=\frac{\sum_{j\in\mathcal{N}_i}w_{ij}z_i^{(j)}}{\sum_{j\in\mathcal{N}_i}w_{ij}}.
\end{equation}

To estimate the reliability of the pose-guided prior, we compute the local trajectory-consistency variance:
\begin{equation}
\sigma_{z,i}^{2}=\frac{\sum_{j\in\mathcal{N}_i}w_{ij}\left(z_i^{(j)}-\hat{z}_i\right)^2}{\sum_{j\in\mathcal{N}_i}w_{ij}}.
\end{equation}
If multiple neighboring poses provide consistent height predictions, the variance is small and the pose-guided prior is reliable. Otherwise, a large variance indicates that the local pose prior may be affected by pose noise, sharp road slope changes, or complex road geometry. Therefore, we define the trajectory-consistency confidence as:
\begin{equation}
q_i=\exp\left(-\frac{\sigma_{z,i}^{2}}{\tau_z^2}\right),
\end{equation}
where $\tau_z$ controls the sensitivity to height inconsistency.

The initial elevation of the Gaussian surfel is set to $\hat{z}_i$. The initial rotation is obtained from the weighted average of neighboring pose rotations. In practice, we use the nearest pose rotation when the local trajectory is smooth, and use the pose-consensus confidence to regularize the subsequent optimization. This strategy extends the original nearest-pose initialization to a more robust multi-pose consensus initialization, reducing the influence of local pose noise without requiring explicit pose covariance.

\subsection{Optimization}
\label{sec:opt}

\subsubsection{RGB and Semantic Rendering}
\label{sec:render}

To render camera-view images, Gaussian surfels are first transformed from world coordinates to camera coordinates using the world-to-camera transformation matrix $\mathbf{W}$, and then projected onto the image plane through a local affine transformation $\mathbf{J}$.  The rendered color can be represented as:
\begin{equation}
\mathbf{c}(\mathbf{p})=\sum_{k=1}^{K}\mathbf{c}_k\alpha_k g_k(\mathbf{p})\prod_{l=1}^{k-1}\left(1-\alpha_l g_l(\mathbf{p})\right),
\end{equation}
where $\mathbf{c}_k$ denotes the color of the $k$-th Gaussian surfel and $\alpha_k$ denotes its opacity. When rendering semantics, the color vector $\mathbf{c}_k$ is replaced by the semantic vector $\mathbf{Sem}_k$. Due to the large number of Gaussian surfels in large-scale road scenes, only Gaussian surfels within a local range in front of the vehicle camera are used for perspective rendering to improve efficiency. For BEV visualization, the road map is rendered in chunks and then stitched together.

In real driving scenarios, different surround-view cameras may have different exposure levels. Therefore, when reconstructing road surfaces from multiple cameras, we introduce learnable exposure parameters $a$ and $b$ for each camera. The final rendered color is formulated as:
\begin{equation}
\mathbf{c}^{\prime}(\mathbf{p})=e^{a}\cdot\mathbf{c}(\mathbf{p})+b.
\end{equation}

\subsubsection{Loss Function}
\label{sec:loss}

For RGB reconstruction, the rendered images are supervised by the camera images. For semantic reconstruction, the rendered semantic maps are supervised by semantic segmentation results inferred by Mask2Former~\cite{cheng2022masked}. A semantic-based road mask $M$ is used to remove non-road elements. The color loss is defined as:
\begin{equation}
L_c=\frac{1}{|M|}\sum_i m_i\cdot|\mathbf{c}_i-\bar{\mathbf{c}}_i|_1,
\end{equation}
where $\bar{\mathbf{c}}_i$ denotes the ground-truth color and $m_i$ is a binary mask value.

The semantic loss is defined as:
\begin{equation}
L_s=\frac{1}{|M|}\sum_i m_i\cdot CE(\mathbf{Sem}_i,\bar{\mathbf{Sem}}_i),
\end{equation}
where $\bar{\mathbf{Sem}}_i$ denotes the semantic pseudo-label and $CE(\cdot)$ denotes the cross-entropy loss. $|M|$ denotes the number of valid pixels in the road mask.

Due to the local smoothness of road surfaces, we introduce an elevation smoothness loss:
\begin{equation}
L_{smooth}=\frac{1}{K}\sum_{i=1}^{N}\sum_{j\in\mathcal{N}(i)}|z_i-z_j|_2^2,
\end{equation}
where $\mathcal{N}(i)$ denotes the $K$ nearest neighboring Gaussian surfels of the $i$-th surfel. Instead of finding nearest neighbors by sorting all surfels according to distance, we assign Gaussian surfels to pixel indexes in the $xy$-plane and query neighboring surfels through the index map, which significantly reduces the computational cost in large-scale scenes.

To preserve the pose-guided road surface prior while reducing the influence of unreliable local pose estimates, we introduce a trajectory-consistency-guided height regularization loss:
\begin{equation}
L_{traj}=\sum_{i=1}^{N}q_i\cdot\rho(z_i-\hat{z}_i),
\end{equation}
where $\hat{z}_i$ is the pose-consensus elevation, $q_i$ is the trajectory-consistency confidence, and $\rho(\cdot)$ is a robust penalty function. When neighboring vehicle poses provide consistent height predictions, $q_i$ becomes large and the pose prior strongly regularizes the surfel elevation. When the local trajectory is inconsistent, $q_i$ becomes small and the optimization relies more on rendering, semantic, smoothness, and optional LiDAR supervision.

Additionally, when LiDAR point clouds are available, they can be used for elevation supervision:
\begin{equation}
L_z=\sum_{i=1}^{N}|z_i-\bar{z}_i|_2^2,
\end{equation}
where $\bar{z}_i$ is obtained by querying the nearest point-cloud elevation in the $xy$-plane. When LiDAR is unavailable, the corresponding loss weight is set to zero.

\section{Experiments}
\begin{table*}[htbp]
  \caption{Results on the nuScenes dataset. The mIoU is expressed as a percentage. ``Elev.'' denotes the RMSE of the elevation error with units in $m$. ``Ep.1'' and ``Ep.2'' denote training one epoch and two epochs respectively. When neither LiDAR point cloud is used, the PSNR of our method is the same as RoMe~\cite{mei2024rome}, and it is improved in both mIoU and Elev. metrics that reflect the 3D structure.  Importantly, we achieve a $\mathbf{53\times}$ speedup. In addition, it can be noticed that there is a significant improvement in all metrics when using LIDAR.}
  \label{tab:metric}
  \centering
  \resizebox{\linewidth}{!}{
  \begin{tabular}{l||ccc|ccc|ccc|ccc|ccc|cccl}
    \toprule
    \multirow{2}{*}{Method} &\multicolumn{3}{c|}{scene-0064} & \multicolumn{3}{c|}{scene-0212}& \multicolumn{3}{c|}{scene-0523}& \multicolumn{3}{c|}{scene-0655}&\multicolumn{3}{c|}{scene-0856}&\multicolumn{4}{c}{Mean}\\
    \cline{2-20}
    & PSNR & mIoU& Elev. & PSNR & mIoU& Elev.& PSNR & mIoU& Elev.& PSNR & mIoU& Elev.& PSNR & mIoU& Elev. & PSNR & mIoU& Elev. &Time\\
    \hline
    \hline
    RoMe~\cite{mei2024rome} & 23.38  &84.48 &0.108 & 25.58&89.11&0.180&25.83&86.37&0.118&20.56&84.24&0.164&22.79&80.53&0.311&23.63&84.95&0.176&1630~\\
    
    Ours w/o LiDAR Ep.1 & 23.46  &84.21 &0.121 & 25.72&91.95&0.105&24.91&87.07&0.121&21.99&86.57&0.101&23.42&77.98&0.285&23.90&85.56&0.147&30.7 ~\textcolor{red}{\textbf{53$\times$}}\\
    Ours w/o LiDAR Ep.2& 23.10  &83.48 &0.125 & 25.15&91.83&0.112&24.86&87.21&0.124&21.74&86.54&0.104&23.03&77.95&0.288&23.57&85.41&0.151&61.1~\textcolor{red}{\textbf{27$\times$}}\\
    \midrule
    Ours w/ LiDAR Ep.1 & 24.58  &92.31 &0.056 & 25.92&93.81&0.070&25.77&91.71&0.054&23.12&92.41&0.063&23.41&82.71&0.232&\textbf{24.56}&90.59&0.095&32.1~\textcolor{red}{\textbf{51$\times$}}\\
    Ours w/ LiDAR Ep.2& 24.44 &93.83 &0.055 & 25.75&94.03&0.068&25.41&92.42&0.052&23.21&93.34&0.062&23.32&84.45&0.221&24.42&\textbf{91.61}&\textbf{0.092}&63.9~\textcolor{red}{\textbf{25$\times$}}\\
    \bottomrule
  \end{tabular}
  }
\end{table*}

\subsection{Dataset}

\textbf{nuScenes:} We conduct experiments on the nuScenes~\cite{caesar2020nuscenes} dataset, which contains 1000 video scenes, each lasting approximately 20 seconds. The data collection vehicle is equipped with 6 surround-view cameras and one LiDAR sensor. The camera and LiDAR frequencies are 12 Hz and 20 Hz, respectively. We use surround-view images, vehicle poses, semantic segmentation results, and optional LiDAR point clouds for road surface mapping. Following previous large-scale road surface reconstruction methods~\cite{mei2024rome, wu2024emie}, we evaluate the proposed method on representative nuScenes scenes covering diverse road structures, illumination conditions, and weather conditions.

\textbf{KITTI:} We also conduct experiments on the KITTI odometry dataset~\cite{geiger2013vision}. KITTI contains 22 sequences collected in urban, residential, and road environments. In particular, sequence 00 covers an area of approximately 600 m $\times$ 600 m, which is suitable for evaluating large-scale road surface mapping. The KITTI vehicle is equipped with stereo color cameras and a LiDAR sensor. In our experiments, we use the left color camera for image supervision and use the LiDAR point clouds for evaluation and optional elevation supervision.

For both datasets, we use a pre-trained Mask2Former~\cite{cheng2022masked} model to obtain semantic segmentation results. To ensure fair comparison with RoMe~\cite{mei2024rome}, we use the same semantic segmentation model and the same semantic categories for evaluation.

\subsection{Experiment Setup}

\textbf{Evaluation Metrics:} Since BEV ground-truth road maps are not directly provided by nuScenes or KITTI, we construct pseudo ground truth using concatenated LiDAR point clouds. Specifically, each LiDAR frame is transformed into the global coordinate system using the vehicle pose. Then, each point is projected onto the nearest camera images, where its color and semantic label are determined by the corresponding projected image pixel. Points that do not belong to road surface elements are removed according to the semantic labels. For points visible in overlapping regions of multiple cameras, the RGB value is computed by averaging the projected colors, and the semantic label is obtained by majority voting.

The accumulated dense LiDAR point clouds are rendered into BEV RGB maps and semantic maps, which are used to evaluate the reconstructed road surface maps. We use Peak Signal-to-Noise Ratio (PSNR) to evaluate RGB reconstruction quality and mean Intersection over Union (mIoU) to evaluate semantic reconstruction accuracy. For geometric evaluation, we use the elevation of the accumulated LiDAR point cloud as the reference. For each reconstructed Gaussian surfel center, we query the nearest LiDAR point within a radius of 0.1 m on the $xy$-plane and compute the elevation error. The root mean square error of the elevation is reported as the elevation RMSE.

\textbf{Implementation Details:} All experiments are conducted on a Linux server with an NVIDIA RTX-4090 GPU. The learning rates for opacity $\alpha$, scale $(s_x,s_y)$, and rotation $\mathbf{R}$ are set to $1\times10^{-4}$. The learning rate of elevation $z$ is proportional to the scene size, initialized with a ratio of $1.6\times10^{-4}$ and gradually decayed to $1.6\times10^{-6}$. The learning rates for color and semantics are set to 0.008 and 0.1, respectively. The learning rate for the exposure parameters is set to 0.001. 

For the proposed adaptive meshgrid, the base BEV resolution is set to 0.05 m/pixel. A grid cell is adaptively refined when its road-structure complexity score is larger than the threshold $\tau_a$. In our implementation, the structure complexity is computed from semantic boundary response, RGB gradient response, and optional elevation variation. The maximum refinement level is set to one for efficiency. For trajectory-consistency-guided pose-robust refinement, we use the neighboring vehicle poses within a local spatial window to estimate the pose-consensus elevation and trajectory-consistency confidence.

To balance reconstruction quality and training speed, we conduct experiments with different BEV resolutions on nuScenes scene-0655. As shown in \ref{fig:resolution}, resolutions greater than or equal to 0.1 m/pixel lead to blurred reconstruction, while resolutions smaller than or equal to 0.02 m/pixel introduce unnecessary computational cost. Therefore, 0.05 m/pixel is selected as the default base resolution.

\begin{figure}[htbp]
  \centering
    \begin{subfigure}{\linewidth} 
    \centering
      \includegraphics[width=\linewidth]{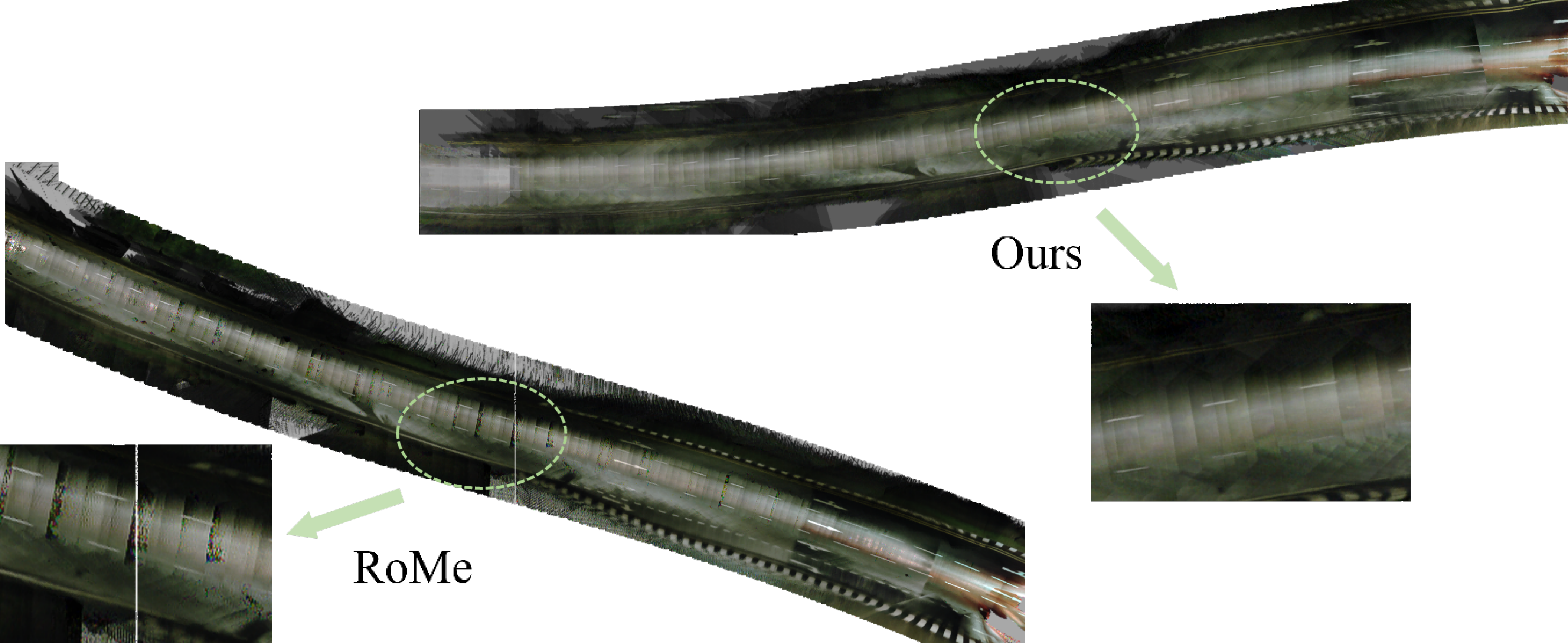}
        \caption{Scene-1079 (Night)}
        \label{fig:1079}
        \vspace{3mm}
    \end{subfigure} 
    \begin{subfigure}{\linewidth}
    \centering
      \includegraphics[width=\linewidth]{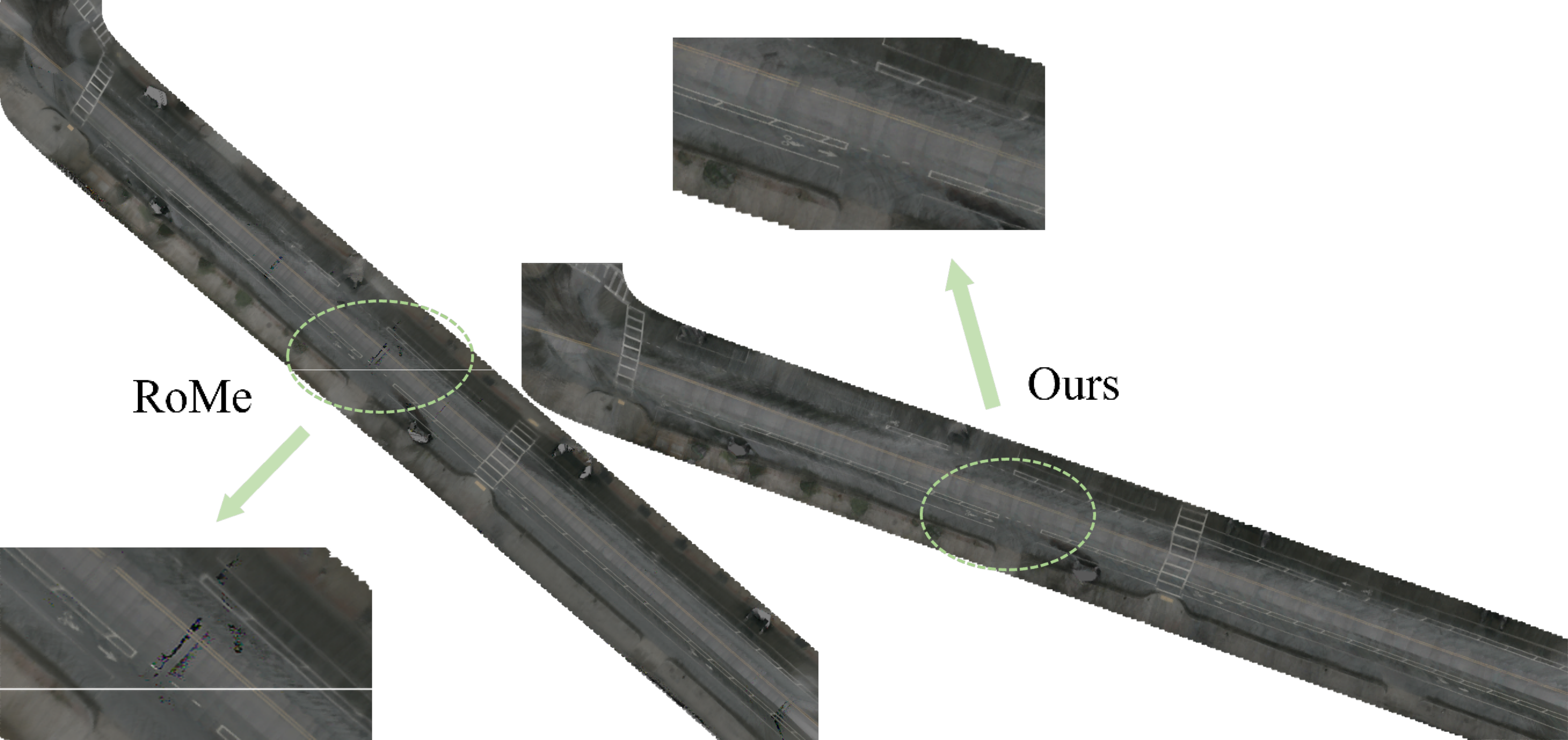}
        \caption{Scene-0580 (Rainy)}
        \label{fig:0580}
    \end{subfigure}
\caption{Comparison of road reconstruction results for night and rainy conditions between RoMe~\cite{mei2024rome} and ours without LiDAR point cloud. In challenging scenes, the reconstruction results of RoMe~\cite{mei2024rome} show obvious RGB noise, and our method is more robust compared to it.}
\label{fig:night_rainy}
\end{figure}

\subsection{Experimental Results}
\label{sec:compare}
The reconstructed road surface maps are rendered in BEV using an orthographic camera. For large-scale scenes, the road surface is rendered in chunks and then stitched into a complete BEV map to improve rendering efficiency.

\textbf{Quantitative Comparison:} Fig.~\ref{tab:metric} reports the quantitative comparison on five nuScenes scenes. We compare the proposed RoGS with RoMe~\cite{mei2024rome}, a representative mesh-based large-scale road surface reconstruction method. Without LiDAR elevation supervision, RoGS achieves comparable RGB reconstruction quality to RoMe in terms of PSNR, while obtaining higher mIoU and lower elevation RMSE. This indicates that the proposed meshgrid Gaussian representation is more effective in reconstructing semantic and geometric road surface structures. When LiDAR point clouds are used for elevation supervision, the reconstruction accuracy is further improved in all metrics, especially in semantic mIoU and elevation RMSE.

In terms of efficiency, the proposed method substantially reduces the optimization time compared with RoMe. With one training epoch, RoGS achieves a significant speedup while maintaining high-quality RGB, semantic, and elevation reconstruction. With two training epochs, semantic reconstruction is slightly improved, while the elevation error remains stable. These results demonstrate that the proposed Gaussian surfel representation and meshgrid layout are more efficient than conventional mesh-based rendering optimization.

\textbf{Qualitative Comparison:} Fig.~\ref{fig:0655_rgb} and \ref{fig:0655_sem} show the reconstructed RGB maps, semantic maps, and elevation maps on nuScenes scene-0655. Compared with RoMe~\cite{mei2024rome}, the proposed method reconstructs lane markings, road boundaries, bicycle markings, and crosswalk regions more clearly. The elevation error maps also show that RoGS produces fewer geometric errors. Without LiDAR elevation supervision, the proposed method can still recover the elevation changes around road edges and zebra crossings to some extent. In contrast, RoMe tends to produce over-smoothed elevation maps because it uses an MLP to model road surface height. When LiDAR supervision is introduced, the reconstructed elevation becomes more consistent with the LiDAR-based reference map.

\textbf{Challenging Conditions:} Fig.~\ref{fig:night_rainy} compares the reconstruction results under challenging night and rainy conditions. Both RoMe and RoGS are evaluated without LiDAR elevation supervision. The results show that RoMe produces obvious RGB noise and less stable road textures under challenging illumination and weather conditions. In contrast, RoGS produces cleaner RGB maps and more stable road structures, demonstrating better robustness in complex real-world driving scenarios.

\begin{figure}[tbp]
  \centering
    \begin{subfigure}{\linewidth} 
    \centering
      \includegraphics[width=\linewidth]{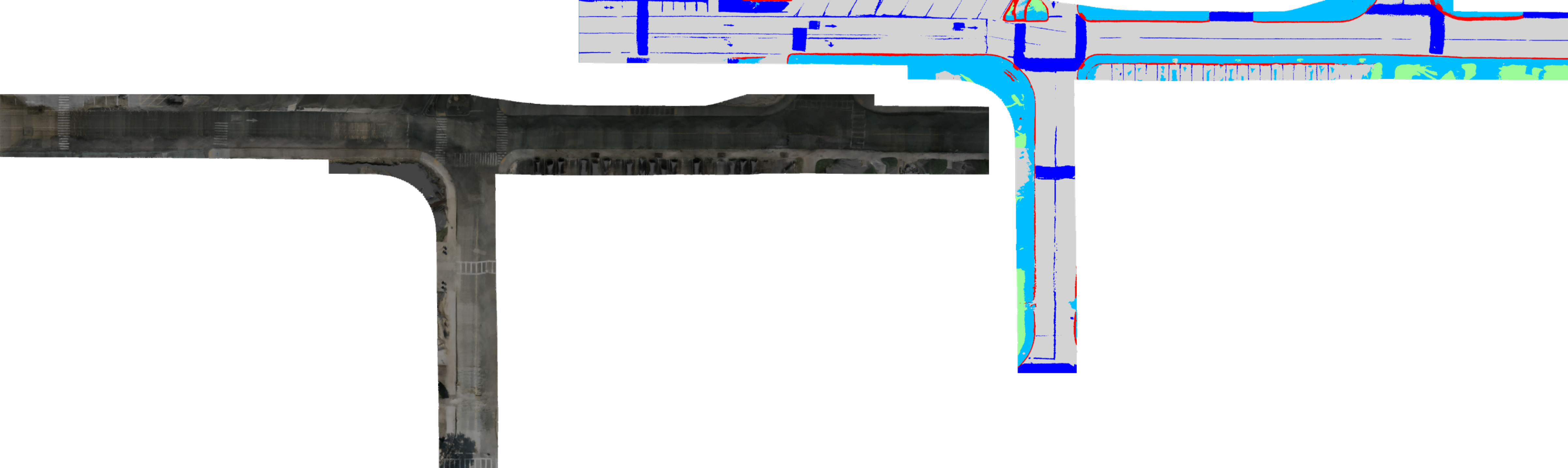}
        \caption{Scene-1}
        \label{fig:scene1}
        \vspace{3mm}
    \end{subfigure} 
    \begin{subfigure}{\linewidth}
    \centering
      \includegraphics[width=\linewidth]{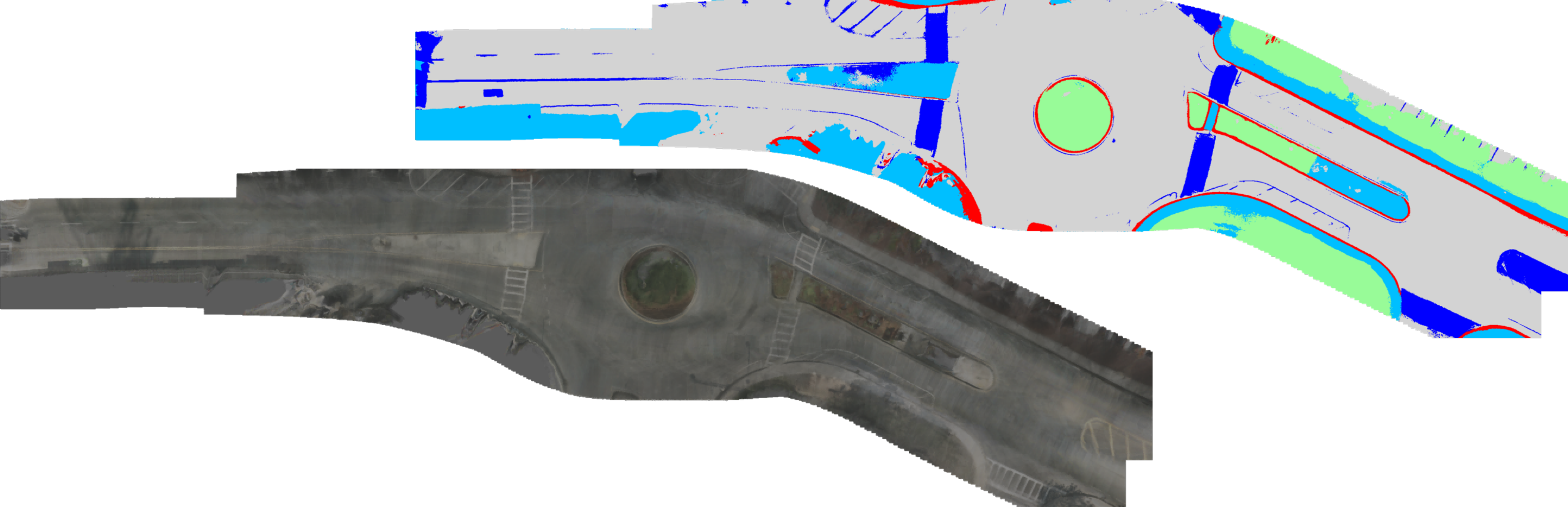}
        \caption{Scene-2}
        \label{fig:scene2}
    \end{subfigure}
\caption{Road surface reconstructions of Scene-1 and Scene-2 with LiDAR point cloud.}
\label{fig:merge_scene}
\end{figure}

\textbf{Large-Scale Multi-Scene Merging:} Fig.~\ref{fig:merge_scene} shows the reconstruction results of merged scenes. Following RoMe~\cite{mei2024rome}, we evaluate two merged scenes, where each merged scene is composed of four individual nuScenes scenes. The proposed method reconstructs complete and visually consistent road surface maps across multiple scenes, demonstrating its scalability for large-scale road surface mapping.

\subsection{Ablation Study}

We conduct ablation studies to analyze the contribution of each component in RoGS. 

\textbf{Effect of Meshgrid Gaussian Representation:} We first evaluate the effectiveness of the proposed meshgrid Gaussian representation. As shown in \ref{tab:ablation}, the meshgrid Gaussian layout achieves better semantic accuracy and lower elevation error than the alternative layout. This is because the adopted meshgrid layout covers the road surface with fewer Gaussian surfels and reduces unnecessary overlap between neighboring surfels. Therefore, it provides a compact and optimization-friendly representation for road surface mapping.

\textbf{Effect of Pose-Based Initialization:} We then evaluate the effect of pose-based initialization. When pose-based initialization is removed, all Gaussian surfels are initialized with $z=0$ and identity rotation. As shown in \ref{tab:ablation}, removing pose-based initialization leads to significant degradation in elevation reconstruction and semantic mapping. This demonstrates that vehicle poses provide an effective geometric prior for road surface initialization.

\textbf{Effect of Adaptive Meshgrid:} To verify the proposed road-structure-aware adaptive meshgrid strategy, we compare the uniform meshgrid and the adaptive meshgrid under similar computational budgets. As shown in \ref{tab:ablation}, the adaptive meshgrid improves the reconstruction quality around traffic-critical regions, including lane markings, road boundaries, curbs, and crosswalks. Compared with the uniform meshgrid, it achieves higher semantic mIoU and lower elevation RMSE while introducing only limited additional Gaussian surfels. These results show that allocating more surfels to structurally complex regions is more effective than uniformly increasing the grid density over the entire road surface.

\textbf{Effect of Trajectory-Consistency-Guided Refinement:} We further evaluate the proposed trajectory-consistency-guided pose-robust refinement. The baseline uses only the nearest vehicle pose to initialize each Gaussian surfel, while our method aggregates height priors from multiple neighboring poses and adaptively weights the pose-guided regularization according to trajectory consistency. As shown in \ref{tab:ablation}, the proposed refinement reduces elevation error and improves semantic reconstruction, especially in regions where the trajectory contains local perturbations or where the road surface changes smoothly along the driving direction. This demonstrates that multi-pose consensus provides a more stable road surface prior than single-pose initialization.

\begin{table}[t]
  \caption{Ablation study of the proposed meshgrid layout, pose initialization,
  adaptive meshgrid, and trajectory-guided refinement.}
  \label{tab:ablation}
  \centering
\scalebox{0.8}{
\setlength{\tabcolsep}{1.0mm}{
  \begin{tabular}{ccccc|ccc}
    \toprule
    Layout-1 & Layout-2 & Pose Init. & Adaptive Meshgrid & Traj. Refine.
    & PSNR$\uparrow$ & mIoU$\uparrow$ & Elev.$\downarrow$ \\
    \midrule
    \checkmark &            &            &            &            
    & 23.33 & 73.28 & 0.966 \\
    
               & \checkmark &   \checkmark         &            &            
    & 23.25 & 81.63 & 0.331 \\
    
              \checkmark &  & \checkmark &            &            
    & 23.38 & 80.47 & 0.326 \\
    
              \checkmark &  & \checkmark & \checkmark &            
    & 24.01 & 85.72 & 0.137 \\
    
       \checkmark        &  & \checkmark & \checkmark & \checkmark 
    & \textbf{24.56} & \textbf{90.59} & \textbf{0.095} \\
    \bottomrule
  \end{tabular}}}
\end{table}

\subsection{Limitations}
\label{limit}
Because the reconstruction of the road surface relies on the accuracy of the pose, it often requires more precise vehicle poses in practical use. Although reconstruction results can be fine-tuned by optimizing road colors and elevations, the capacity for this adjustment is limited.  Furthermore, due to the sparse texture information on the road surface, relying solely on ground color and semantics to accurately restore road height can be quite challenging. This issue is more pronounced in RoMe~\cite{mei2024rome}. Therefore, introducing point clouds for elevation supervision is a highly effective method. For vehicles without equipped LiDAR, point clouds obtained from COLMAP~\cite{Schonberger2016structure} can be used as an alternative.

\section{Conclusion}

In this paper, we propose RoGS, a robust and efficient framework for large-scale road surface mapping with adaptive meshgrid Gaussian representation. By modeling road surfaces with mesh-distributed 2D Gaussian surfels, RoGS explicitly represents RGB, semantic, and elevation information while conforming to the thin-surface property of roads. To improve representation efficiency and preserve traffic-critical structures, we introduce a road-structure-aware adaptive meshgrid strategy, which allocates denser Gaussian surfels to complex regions such as lane markings, road boundaries, curbs, and height discontinuities. In addition, we propose a trajectory-consistency-guided pose-robust refinement strategy, which aggregates local surface priors from multiple neighboring vehicle poses and adaptively reduces the influence of unreliable pose guidance. Extensive experiments on KITTI and nuScenes demonstrate that RoGS achieves high-quality road surface reconstruction in various challenging real-world scenes while maintaining high optimization efficiency. These results show the potential of meshgrid Gaussian representation for scalable road surface mapping and autonomous driving applications.

\bibliographystyle{unsrt}  
\bibliography{neurips_2025} 

\end{document}